\documentclass{article}

\usepackage[preprint]{neurips_2023}

\usepackage{natbib}
\setcitestyle{numbers,square}

\usepackage[utf8]{inputenc} 
\usepackage[T1]{fontenc}    
\usepackage{hyperref}       
\usepackage{url}            
\usepackage{booktabs}       
\usepackage{amsfonts}       
\usepackage{nicefrac}       
\usepackage{microtype}      
\usepackage{xcolor}         
\usepackage{multicol, multirow}
\usepackage{graphicx}
\usepackage{makecell}

\title{Flexibly Scaling Large Language Models' Contexts Through Extensible Tokenization}

\author{%
  Ninglu Shao$^{1,2}$\thanks{Ninglu Shao (ninglu\_shao@ruc.edu.cn), Shitao Xiao, and Zheng Liu are the co-first authors.}
  ~~
  Shitao Xiao$^{1}$
  ~~
  Zheng Liu$^{1}$\thanks{Zheng Liu (zhengliu1026@gmail.com) is the correponding author.}
  ~~
  Peitian Zhang$^{1,2}$ \\
  1: Beijing Academy of Artificial Intelligence, \\
  2: Gaoling School of Artificial Intelligence, Renmin University\\ 
}

\begin{document}

\maketitle





\begin{abstract}
Large language models (LLMs) are in need of sufficient contexts to handle many critical applications, such as retrieval-augmented generation and few-shot learning. However, due to the constrained window size, the LLMs can only access to the information within a limited context. Although the size of context window can be extended by fine-tuning, it will result in a substantial cost in both training and inference stage. In this paper, we present \textbf{Extensible Tokenization} as an alternative method which realizes the flexible scaling of LLMs' context. Extensible Tokenization stands as a midware in between of the tokenized context and the LLM, which transforms the raw token embeddings into the extensible embeddings. Such embeddings provide a more compact representation for the long context, on top of which the LLM is able to {perceive more information} with the same context window. Extensible Tokenization is also featured by its {flexibility}: the scaling factor can be flexibly determined within a feasible scope, leading to the extension of an arbitrary context length at the inference time. Besides, Extensible Tokenization is introduced as a {drop-in component}, which can be seamlessly plugged into not only the LLM itself and but also its fine-tuned derivatives, bringing in the extended contextual information while fully preserving the LLM's existing capabilities. We perform comprehensive experiments on long-context language modeling and understanding tasks, which verify Extensible Tokenization as an effective, efficient, flexible, and compatible method to extend LLM's context. Our model and source code will be made publicly available at https://github.com/FlagOpen/FlagEmbedding.

\end{abstract}

\begin{figure}[htb]
    \centering
    \includegraphics[width=\textwidth]{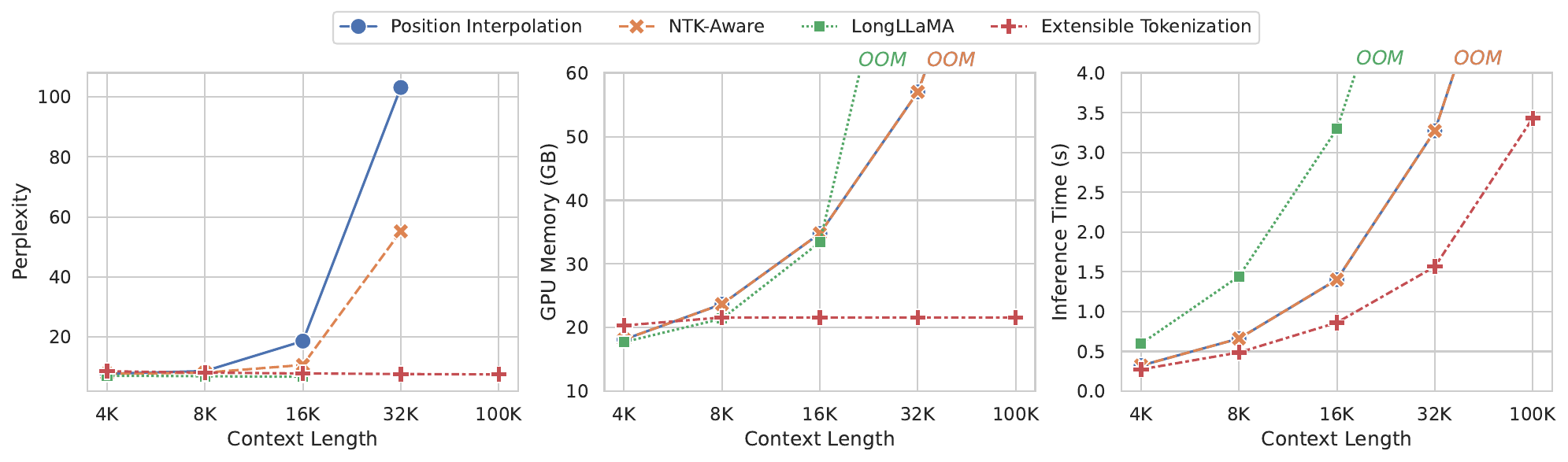}
    \caption{Comparison between Extensible Tokenization and other context extension methods, including 1) Position Interpolation \cite{chen2023extending}, 2) NTK-Aware Scaled RoPE \cite{ntkaware2023}, 3) LongLLaMA \cite{tworkowski2023focused}. 
    Extensible Tokenization presents a superior long-context language modeling capability, along with better efficency in terms of memory and time. 
    PPL is measured on PG19 \cite{raecompressive2019} following the method in \cite{chevalier2023adapting}}
    \label{fig:cmp_3}
\end{figure}

\section{Introduction}
Large language models (LLMs) need to process long-sequence data in order to accomplish many critical tasks, like retrieval augmented generation and in-context learning. Unfortunately, the existing LLMs are constrained by the sizes of their context windows, which are far from enough to fully cover the input data in the corresponding scenarios. Although the LLM's context window can be extended by fine-tuning \cite{chen2023longlora,longchat2023,peng2023yarn}, it will lead to a considerable cost at both training and inference time. Besides, the continual fine-tuning over long-sequence data is likely to impair the LLM's original capability on shorter contexts, which is unfavorable to its practical usage. Alternatively, the LLMs can be also modified by many other techniques to establish longer context windows, such as sparse attention~\cite{child2019sparse_transformers,beltagy2020longformer,zaheer2020bigbird,ding2023longnet}, stream processing~\cite{xiao2023streamingllm,han2023lm_infinite}, compressed memory~\cite{aydar2023rmt,chevalier2023autocompressors,wu2023gist,rae2020compressive,huang2023selective_cache}, etc. However, the existing methods are still limited by manifold problems. For example, the sparse attention calls for customized GPU kernels which are not supported by the standard infrastructures; the stream processing ignores the information beyond the context window instead of making effective use of it; the compressed memory are still limited by the substantial loss of contextual information and the incompatibility with the existing LLMs.  

In fact, the LLM's context capacity, i.e. the contextual information that the LLM is able to perceive, is jointly determined by two critical factors. One is the \textit{size of context window}, which regulates the amount of input units the LLM can intake. The other one is the \textit{information density}, which reflects the information encoded by each input unit. Currently, the typical input unit to LLM is token embedding, which is relatively information sparse because it only encodes the information about each individual token. Once with more compact and informative input units, the LLM will be able to perceive more information with the same context window. 

Based on this fundamental principle, we propose \textbf{Extensible Tokenization} as a new method to extend the LLM's context. Extensible Tokenization stands as a midware in between of the input context and the LLM, which transforms raw token embeddings into \textit{extensible embeddings} for the compact representation of the input context. Extensible Tokenization shares a common high-level spirit as the previous methods on context compression \cite{chevalier2023adapting,wu2023gist,chevalier2023autocompressors,dai2020funnel}. However, it enjoys a substantially improved performance on context extension, a high flexibility of usage, and a high compatibility with the downstream LLM thanks to a series of characters about its architecture and training. 


Instead of modifying the original model architecture, we employ another pre-trained language model as the stand-alone module for Extensible Tokenization. It transforms the raw input into its output embeddings, and then performs down-scaling by the scaling factor $k$ (e.g., $k=32$) for the extensible embeddings. Given the dramatic scale of compression, the learning process must be properly designed such that the contextual information can be fully preserved. In our work, we propose to learn the extensible embeddings through auto-regression (AR), where the compressed context can well assist the downstream LLM for the prediction of the next tokens. Besides, we further optimize the sample efficiency of training through the two-stream processing. With merely two passes of feed-forward, it will let the training loss to be derived from every single token of one training sample. By minimizing the comprehensive prediction losses from the data, the extensible embeddings can be effectively learned as compact but equally informative representations of the context.




Extensible Tokenization presents a highly flexible solution to extend the LLM's context. In particular, the scaling factor $k$ can be freely specified at the inference time, leading to the flexible extension for diverse context lengths. 
Besides, the extensible embeddings and the raw token embeddings can be seamlessly blended with each other, with the extensible embeddings applied for the verbose part of the input (e.g., the retrieved documents) and the raw token embeddings used for the concise part of the input (e.g., questions or instructions). The collaborative play of both types of embeddings contributes to a superior quality for the LLM's utilization of the extended context. 


The extensible tokenizer is learned with the downstream LLM's parameters fixed all the time. Therefore, it can work as a compatible plug-and-play component, introducing the extended context as extensible embeddings without compromising the LLM's original performance with the raw token embeddings. Notably, beyond the LLM where the training is conducted, it is empirically observed that the extensible tokenizer also exhibits a strong compatibility with various fine-tuned derivatives of the LLM. Such a property indicates the potential of Extensible Tokenization as a versatile module for the extension of the context across a family of closely related LLMs.



With Extensible Tokenization, the inference can be streamingly conducted by sessions. In each session, a new token is predicted based on the extensible embeddings from previous sessions and the preceding raw tokens in the same session. The stream processing leads to a linear time complexity and a substantial reduction of memory usage, which benefits the efficiency of processing long-sequence data. For many real-world scenarios, e.g., retrieval-augmented generation, the extensible embeddings can be pre-computed and cached, which will further save the computation cost of online inference. 



We apply the first 8 layers of LLaMA-2-7B (chat) model \cite{touvron2023llama-b} as the extensible tokenizer's backbone, where it is used for the context extension for another downstream LLaMA-2-7B (chat) model. The training is performed on top of the sampled data from RedPajama \cite{together2023redpajama} and LongAlpaca \cite{chen2023longlora}. So far, Extensible Tokenization is trained with a maximum scaling factor of 32, enabling the context length of LLaMA-2-7B to be extended over 100K. In our experiment, Extensible Tokenization achieves superior performances on both long-context language modeling and understanding tasks. Besides, the context extension capability can be well transferred to the fine-tuned derivatives of LLaMA-2-7B, like LongChat \cite{longchat2023} and LongAlpaca \cite{chen2023longlora}. Therefore, the Extensible Tokenization is verified as an effective, efficient, flexible, and compatible method for the extension of LLM's context.

\begin{figure}[t]
    \centering
    \includegraphics[width=\textwidth]{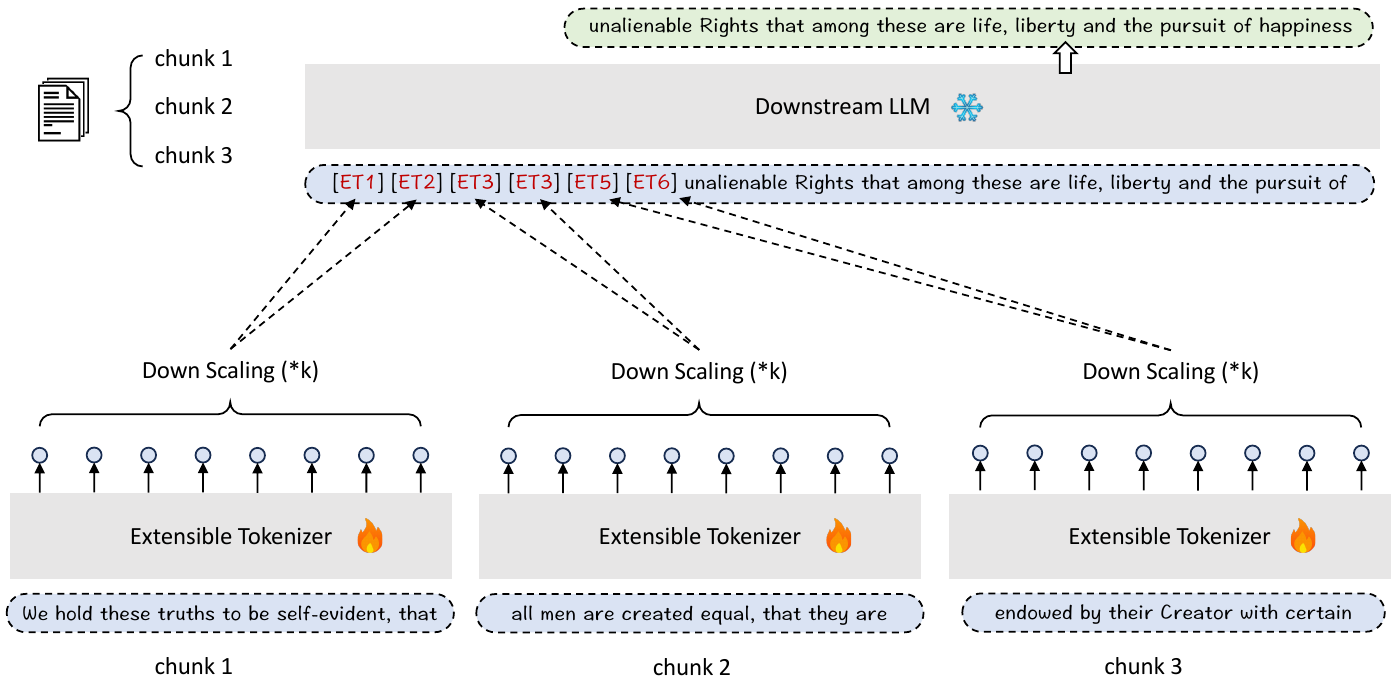}
    \caption{Extensible Tokenization. The input data is chunked into equal-sized sub-sequences. Each sub-sequence is transformed and compressed as the extensible embeddings. The new tokens are predicted based on the extensible embeddings from the preceding chunks and the token embeddings in the same chunk. The extensible tokenizer is learned with a fixed downstream LLM.}
    \label{fig:supertoken}
\end{figure}

\section{Extensible Tokenization} 

\subsection{Framework}
The workflow of Extensible Tokenization is shown as Figure \ref{fig:supertoken}. For each long-sequence input $X$, we perform the following three steps which enables the long context to be utilized by the LLM. Firstly, the input $X$ is chunked into sub-sequences: $\{X_{1}, ... X_{N}\}$.  The sequence length of each chunk $L_i$ is set to be the maximum window size of the extensible tokenizer, e.g., $L_i = 4096$ with LLaMA-2, which will best preserve the coherence of the chunking result. Secondly, the sub-sequence of each chunk is transformed by the extensible tokenizer into the output embeddings. The output embeddings are down-scaled by the scaling factor $k$ (e.g., $k=16$ or $32$), where $L/k$ extensible embeddings (denoted as ET) are generated as the condensed representation of the raw input. Finally, the new tokens are predicted conditioned on the extensible embeddings from the preceding chunks and the raw token embeddings within the recent context. 

\subsection{Extensible Embedding}

As introduced, the raw token embedding, which is merely corresponding to one individual token, is information sparse. In contrast, the extensible embedding is presented to serve as a highly compact but equally informative representation of the context. For this purpose, we employ another language model as the extensible tokenizer (denoted as $\mathrm{LM}_{et}$), which transforms the raw input of each sub-sequence $X_i: \{x_{i,1}, ... x_{i,L}\}$ into the sequence of output embeddings $O_i$: 
\begin{equation}
    O_i: \{o_{i,1}, ... o_{i,L}\} \leftarrow \mathrm{LM}_{et}(x_{i,1}, ... x_{i,L}; \theta_{et}).
\end{equation}
On top of an expressive language model, the rich contextual information within $X_{i,:l}$ can be encoded by the corresponding output embedding $o_{i,l}$. The output embeddings are further down-scaled by the scaling factor $k$, where $m$ ($m = L/k$) extensible embeddings ($et_{i,*}$) are generated for $X_i$: 
\begin{equation}
    \{ et_{i,1}, ..., et_{i,m} \} \leftarrow \mathrm{DownScale}(\{o_{i,1}, ... o_{i,L}\}). 
\end{equation}
There can be many alternative ways to realize the functionality of down-scaling, e.g., with any pooling functions along the sequence dimension. In our work, we simply down-scale the output embeddings by through strided sampling, where the last embedding in every $k$ steps is chosen, i.e., $et_{i,j} \leftarrow o_{i,k{\times}j}$. Despite simplicity, such a realization is empirically effective and leads to a high flexibility of usage.

\begin{figure}[t]
    \centering
    \includegraphics[width=\textwidth]{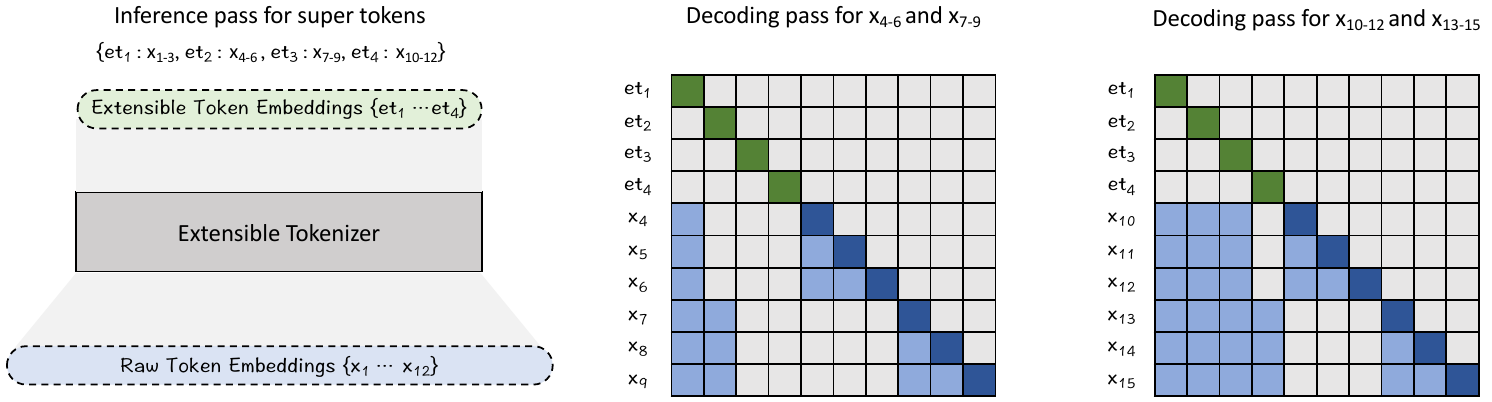}
    \caption{Two-Stream AR. In the first pass, the raw token embeddings are transformed into extensible embeddings (with the scaling factor $k=3$). In the second pass (given a window size of 10), the auto-regression is accomplished in two steps, with the $x_{1-3}$ and $x_{4-6}$ predicted in the first step, and $x_{7-9}$ and $x_{10-12}$ predicted in the second step.} 
    \label{fig:supertoken_attention}
\end{figure}

\subsection{Two-Stream AR}


Extensible Tokenization can be learned by auto-regression (AR), where the loss is minimized for the prediction of next tokens conditioned on the extensible embeddings from the preceding context. Although the auto-regression tasks can be simply performed by having the long context transformed into extensible embeddings and predicting the last few tokens within a training instance (e.g., predicting the answer to a question based on the extensible embeddings of a long document), it will severely restrict the training effect because the long context accounts for the majority of computation cost whereas no prediction loss can be derived from it.

In our work, we propose the two-stream AR to optimize the sample efficiency of training (Figure \ref{fig:supertoken_attention}). In the first pass of inference, the extensible embeddings are generated for the entire context. For example, with a chunk size of 3 and an scaling factor of 3, the input data $X=\{x_1, ... x_{15}\}$ is transformed into the extensible embeddings $\{et_{1,1}, et_{2,1}, et_{3,1}, et_{4,1}\}$ (the last chunk is exempted). In the second past, each single token within the long context is streamingly predicted by chunks. Particularly, the prediction is made conditioned on the extensible embeddings from the previous chunks and the preceding raw token embeddings within the same chunk. Formally, 
\begin{equation}
    \min\limits_{\theta_{et}} \sum\nolimits_{X}\sum\nolimits_{i>1} \log P(x_{i,j} | et_{1,1}, ... et_{i-1,k}, x_{i,1}, ... x_{i,j-1} | \theta, \theta_{et}). 
\end{equation}
For example, $x_6$ is predicted based on $st_{1}$ (representing $x_{1-3}$) and $x_4$. Note that the chunk size of training is made much smaller than the LLM's window size (e.g., 512), where the prediction of new tokens can mostly rely on the contextual information offered by the extensible embeddings. Thanks to the above processing, the prediction loss can be comprehensively derived from the each training instance, enabling the extensible tokenizer to be effectively learned from general long-context data, such as RedPajama \cite{together2023redpajama} and LongAlpaca \cite{chen2023longlora}. We also randomly sample the extension ratio $k$ from a candidate scope (e.g., [2, 4, 8, 16, 32]) for each training instance, which helps the model to generalize for the extension of diverse context lengths. 

When the extensible tokenizer is learned, the downstream LLM's parameters ($\theta$) are always fixed. Consequently, the extensible tokenizer can work as a compatible drop-in component to the downstream LLM, bringing in new information from the extended context without affecting the LLM's performance with the raw token embeddings. Besides, we also empirically find that the extensible tokenizer can maintain a high compatibility with many full-parameter fine-tuned derivatives of the downstream LLM. Such a property makes it a general module to extend the context length for a family of closely related LLMs. 


\section{Inference}


The inference process with the Extensible Tokenization can be divided into the online and offline scenario. Particularly, the online usage deals with the scenario where the long-sequence data is streamingly presented (e.g., conversation). In this scenario, the generation process is conducted in consecutive sessions. In each session, the downstream LLM predicts the new tokens based on the extensible embeddings from the previous sessions and the raw token embeddings within the current session. The current session comes to its end when the total sum of both types of embeddings (denoted as $N_{et}$ and $N_{raw}$, respectively) reaches the maximum window size of the downstream LLM ($L^*$): $N_{et} + N_{raw} = L^*$. The newly generated tokens will be condensed as the extensible embeddings at the end of the current session ($N_{et} \leftarrow N_{et} + N_{raw}/k$), where the new session can be conditioned on the augmented extensible embeddings. The session-based workflow is free from processing the long context directly, which will preserve a small memory footprint. Besides, the inference time will become linear to the context length, which will benefit the processing of long-sequence data. 

As for the offline scenario, the long-sequence data is fully presented in advance (e.g., the documents for RAG and reading-compression). As a result, the extensible embeddings can be pre-computed for the entire data, which will lead to a more competitive efficiency at the inference time. In fact, it is OK to simply save the entire output embeddings in the offline stage, and flexibly sample for the extensible embeddings at the inference time based on the concrete scaling factor.

\section{Experiments}

In this section, we conduct experimental studies to investigate the following key issues about Extensible Tokenization. 1) The effectiveness of context extension. 2) The flexibility and compatibility. 3) The running efficiency. 4) The influential factors about the empirical performance. 

\subsection{Experimental Settings}
In our experiment, we make use of the LLaMA-2-7B (chat) model \cite{touvron2023llama} as our downstream LLM. By default, we take the first 8 layers of LLaMA-2-7B (chat) as the initialized backbone for the extensible tokenizer. The training process takes place on a single Nvidia 8×A800 GPU machine, with a batch size of 8 and a learning rate of $5e^{-5}$ using the linear scheduler. The training is consecutively performed with 90K sampled instances from Redpajama \cite{together2023redpajama} and 10K training instances collected by LongAlpaca \cite{chen2023longlora}. As introduced, the extensible tokenizer is trained while the downstream LLM's parameters are always fixed. 

We consider the following baselines in our experiment. 1) The context extension method without fine-tuning, including Positional Interpolation (PI) \cite{chen2023extending}, the NTK-Aware Scaled RoPE (NTK) \cite{ntkaware2023}, and StreamingLLM (Stream) \cite{xiao2023efficient}. 2) The finetuned full-attention method, including LongAlpaca-7B-16K \cite{chen2023longlora} and LongChat-7B-32K \cite{longchat2023}. 3) The finetuned method with extra architectures to handle long contexts, including AutoCompressor-7B-6K \cite{raecompressive2019} and LongLlama \cite{tworkowski2023focused}. All of the baselines are based on LLaMA-2-7B, except LongLLaMA which leverages CodeLLaMA \cite{roziere2023code}. 

\subsection{Main Results}



\begin{table*}
    \footnotesize
    \centering
    \begin{tabular}{l|ccccc|ccccc}
    \toprule
        \multirow{2}{*}{\textbf{Model}} & \multicolumn{5}{c|}{PG19} & \multicolumn{5}{c}{Books3} \\
        & \textbf{4K} & \textbf{8K} & \textbf{16K} & \textbf{32K} & \textbf{100K} & \textbf{4K} & \textbf{8K} & \textbf{16K} & \textbf{32K} & \textbf{100K} \\
    \midrule
        LLaMA-2-7B & 7.77 & >$10^3$ & >$10^3$ & >$10^3$ & OOM & 4.21 & >$10^3$ & >$10^3$ & >$10^3$ & OOM \\
        PI & 7.77 & 8.68 & 18.65 & >$10^2$ & OOM & 4.21 & 5.99 & 11.4 & 69.8 & OOM \\
        NTK & 7.77 & 8.13 & 10.71 & 55.22 & OOM & 4.21 & 5.10 & 7.71 & 52.3 & OOM \\
        Streaming & 7.98 & 8.01 & 8.00 & 8.00 & 8.00 & 4.32 & 4.34 & 4.33 & 4.33 & 4.34 \\
    \midrule
        LongAlpaca-16K & 8.45 & 8.15 & 8.12 & >$10^3$ & OOM & 4.93 & 4.67 & 4.64 & >$10^3$ & OOM \\
        LongChat-32K & 7.59 & 7.25 & 7.00 & 6.85 & OOM & 4.12 & 3.95 & 3.87 & 3.85 & OOM \\
        AutoCompressor-6K & 26.9 & >$10^3$ & $10^3$ & >$10^4$ & OOM & 17.1 & >$10^3$ & >$10^3$ & >$10^4$ & OOM \\
        LongLLaMA-32K & 7.12 & 6.95 & 6.78 & OOM & OOM & 3.99 & 3.90 & 3.84 & OOM & OOM \\
    \midrule
        ExtenToken ($\times16$) & 7.75 & 7.48 & 7.38 & 7.31 & >$10^2$ & 4.32 & 4.20 & 4.15 & 4.13 & >$10^3$ \\
        ExtenToken ($\times32$) & 8.61 & 8.15 & 7.87 & 7.69 & 7.54 & 4.67 & 4.48 & 4.36 & 4.28 & 4.25 \\
    \bottomrule
    \end{tabular}
    \caption{Language modeling performance (measured by perplexity) on PG19 and Books3.}
    \label{tab:ppl}
\end{table*}

\begin{figure}[htb]
    \centering
    \includegraphics[width=\textwidth]{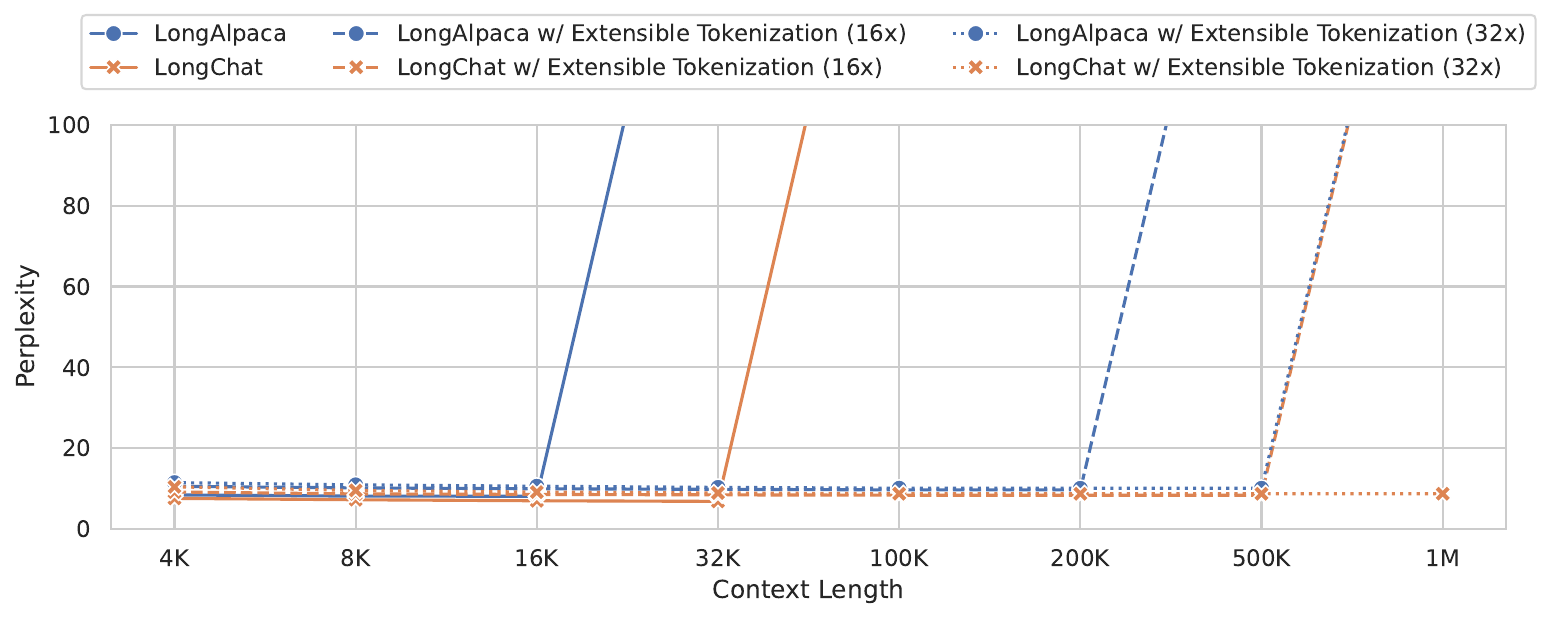}
    \caption{The extensible tokenizer trained on LLaMA-2-7B can be directly utilized by LongAlpaca-16K and LongChat-32K, leading to further scaling of their context lengths by $\times16$ and $\times32$ (with PPL measured on PG19). Remarkable, the context length of LongChat can be extended to 1 million.} 
    \label{fig:ppl_compatibility}
\end{figure}

\subsubsection{Long-Context Language Modeling}
The long-context language modeling is evaluated with PG19 \cite{raecompressive2019} and Books3 \cite{pile} dataset. Following the method used by Alexis et al. \cite{chevalier2023adapting}, 
the PPL is measured by predicting the last 512 tokens based on the preceding context. There are two alternative settings about the Extensible Tokenization, denoted as ExtenToken ($\times16$) and ExtenToken ($\times32$), where the scaling factor is set as 16 and 32, respectively. The evaluation results are reported in Table \ref{tab:ppl}, where the following observations can be derived. 

Firstly, with the extended context length, Extensible Tokenization leads to a notable advantage over LLaMA-2-7B. Besides, its relative improvement is more significant than the fine-tuning free methods, while being comparable to the finetuned baselines. Such an observation indicates that the extended contextual information introduced by Extensible Tokenization can be effectively utilized by the LLM. Secondly, by switching to larger scaling factors, Extensible Tokenization is able to flexibly support longer expansions of the context length than other baselines. Particularly, by switching the scaling factor from $16$ to $32$, LLaMA-2-7B's context length can be continually extended up to $128K$ ($32 \times 4K$). The extended contextual information further improves the performance of language modeling. 
Thirdly, the well-trained extensible tokenizer for LLaMA-2-7B preserves a high compatibility to its fine-tuned derivatives, as it can be directly applied for the context extension of LongAlpaca and LongChat (Figure \ref{fig:ppl_compatibility}). In particular, the context lengths of the two models can also be effectively extended on top of Extensible Tokenization, with LongChat-32K's context length going beyond one million tokens remarkably ($32 \times 32K$).


\subsubsection{Long-Context Understanding}

We perform additional evaluations using three long-context understanding tasks from LongBench \cite{bai2023longbench}, including single-doc QA, multi-doc QA, and summarization. For each evaluation sample, the scaling factor is adjusted case-by-case, which ensures the input data can just fit into the context window of the corresponding LLM (e.g., 4K with LLaMA-2-7B). The evaluation results are presented in Table \ref{tab:longbench}, showing that Extensible Tokenization (denoted as Llama-2-7B w. ExtenToken) substantially improves upon the LLaMA-2-7B baseline for both single-doc QA and multi-doc QA tasks. Interestingly, the observation differs for summarization, where the extended context contributes little to the empirical performance (so is the case with LongAlpaca and LongChat where the improvements are mainly resulted from fine-tuning rather than the extended context). The above improvements are pronounced, considering that Extensible Tokenization works with a shorter context length (resulting in much less GPU memory usage and inference time) and does not impact the LLM's original parameters. 

Similar with our previous exploration, we directly apply the well-trained extensible tokenizer from LLaMA-2-7B for LongAlpaca and LongChat (denoted as LongAlphaca w. ExtenToken and LongChat w. ExtenToken, respectively). Because the majority of the evaluation samples' sequence lengths can be fully covered by the two models (whose context lengths are $16K$ and $32K$, respectively), the extensible tokenizer is introduced mainly for the compression of the data instead of introducing extra information. In other words, LongAlpaca and LongChat can make use of much shorter inputs (within 4K) for the completion of their tasks. Notably, the extensible tokenizer exhibits a strong compatibility with the two fine-tuned derivatives of LLaMA-2-7B: for both single-doc QA and multi-doc QA, it effectively preserves the two models' full-scale performances with the highly compressed inputs.



\begin{table*}[t]
    \centering
    \small
    \begin{tabular}{l|c|ccc}
    \toprule
        Model & Length & Single-Doc QA & Multi-Doc QA & Summarization \\
    \midrule
        Llama-2-7B & 4k & 24.90 & 22.60 & 24.70 \\
        Llama-2-7B w. PI & 16k & 18.98 & 17.16 & 25.03 \\
        Llama-2-7B w. NTK & 16k & 23.21 & 23.34 & 24.40 \\
        Llama-2-7B w. Stream & 16k & 21.47 & 22.22 & 22.20 \\
    \midrule
        Llama-2-7B w. ExtenToken* & 4k & 25.56 & 26.92 & 24.63 \\
    \midrule
    \midrule
        LongAlpaca-16K (4k) & 4k & 26.81 & 24.44 & 26.93 \\
        LongAlpaca-16K (8k) & 8k & 28.61 & 24.83 & 27.91 \\
        LongAlpaca-16K (16k) & 16k & 28.36 & 28.16 & 27.77 \\
    \midrule    
        LongAlpaca-16K w. ExtenToken* & 4k & 28.61 & 28.32 & 26.88 \\
    \midrule
    \midrule
        LongChat-32K (4k) & 4k & 28.14 & 21.88 & 26.59 \\
        LongChat-32K (8k) & 8k & 29.39 & 21.69 & 27.03 \\
        LongChat-32K (16k) & 16k & 30.85 & 23.33 & 26.79 \\
        LongChat-32K (32k) & 32k & 30.98 & 23.96 & 26.82 \\
    \midrule
        LongChat-32K w. ExtenToken* & 4k & 30.12 & 23.51 & 25.91 \\ 
    \bottomrule
    \end{tabular}
    \caption{The evaluation of long-context understanding tasks from LongBench.} 
    \label{tab:longbench}
\end{table*}


\begin{table*}[t]
    \small
    \centering
    \begin{tabular}{l|ccccc|ccccc}
    \toprule
        \multirow{2}{*}{\textbf{Model}} & \multicolumn{5}{c|}{\textbf{GPU Memory (GB)}} & \multicolumn{5}{c}{\textbf{Inference Time (s)}} \\
        & 4K & 8K & 16K & 32K & 100K & 4K & 8K & 16K & 32K & 100K \\
    \midrule
        LongChat-32K & 18.12 & 23.68 & 34.79 & 57.03 & OOM & 0.32 & 0.65 & 1.43 & 3.32 & OOM \\
        StreamingLLM
        & 15.11 & 15.11 & 15.11 & 15.11 & 15.11 & - & - & - & - & - \\
        LongLLaMA & 17.73 & 21.40 & 33.41 & OOM & OOM & 0.60 & 1.44 & 3.30 & OOM & OOM \\
    \midrule
        ExtenToken (on) & 20.33 &21.59 & 21.59 & 21.59 & 21.59 & 0.28 & 0.49 & 0.86 & 1.57 & 3.43 \\
        ExtenToken (off) & 13.96 & 14.21 & 14.75 & 15.79 & 17.54 & 0.08 & 0.08 & 0.10 & 0.12 & 0.23 \\
    \bottomrule
    \end{tabular}
    \caption{Efficiency analysis (FlashAttention-2 is enabled for LongChat).}
    \label{tab:memory&time}
\end{table*}


\subsubsection{Efficiency Analysis}
We make analysis for the running efficiency in terms of GPU memory usage and inference time. The performance is measured by taking the average value of 100 forward passes where the last 512 tokens are predicted based on the preceding context. All the experiments are based on one single Nvidia A800-80G GPU. LongChat is the full-attention method, where the FlashAttention-2 is enabled \cite{dao2023flash}. StreamingLLM relies on stream processing, whose window size is set to 2048. It is exempted from time evaluation because its current stepwise implementation is too slow. We consider the two alternative ways of inference with Extensible Tokenization ($k=32$): the online processing (on) where the input context is streaming presented, and the offline processing (off) where the input context is presented in advance (where the extensible embeddings can be pre-computed). The following observations can be derived from the evaluation results in Table \ref{tab:memory&time}. 

First of all, Extensible Tokenization leads to a very economic usage of GPU memory, which is much smaller than the full-attention methods when the input sequence is long. As introduced, the memory usage of ExtenToken comes from two sources. One is the generation of extensible embeddings, which is performed by sessions with sequence length no more than 4K (ExtenToken (off) is exempted from this step, thus taking even less GPU memory). The other one is the final inference stage based on the extensible embeddings, where the sequence length is much shorter than the raw input. As a consequence, Extensible Tokenization is free from processing the entire long sequence simultaneously, which substantially reduces the memory cost and ensures the extension for a super long context. 

Secondly, Extensible Tokenization exhibits a linear growth of the inference time, as the majority of computation is spent on the session-based generation of extensible embeddings. Besides, with the extensible embeddings pre-computed during the offline stage, the inference time of ExtenToken (off) substantially outperforms other methods. This property suggests its potential value to scenarios, like retrieval augmented generation, where the long-context data can be presented in advance. 


\begin{table*}[t]
    \centering
    \small
    \begin{tabular}{l|l|cccc|c}
    \toprule
        \textbf{Factor} & \textbf{Setting} & \textbf{4K} & \textbf{8K} & \textbf{16K} & \textbf{32k} & \textbf{Single-doc QA} \\
    \midrule
        \multirow{3}{*}{Down scaling method} 
        & Random down-sampling & 8.22 & 7.86 & 7.86 & 7.64 & 23.39 \\
        & Terminal down-sampling & 8.23 & 7.88 & 7.66 & 7.58 & 24.04 \\
        & Strided down-sampling* & 7.75 & 7.48 & 7.38 & 7.31 & 25.56 \\ 
    \midrule
        \multirow{2}{*}{Scaling factor ($k$)} 
        & Monotonous ($k = 16$) & 7.55 & 7.40 & 7.32 & 7.29 & 21.37 \\
        & Dynamic Sampling* & 7.75 & 7.48 & 7.38 & 7.31 & 25.56 \\         
    \midrule
        \multirow{3}{*}{Extensible tokenizer size} 
        & First 4-layer (Llama-2-7B) & 7.89 & 7.64 & 7.52 & 7.46 & 23.32 \\
        & First 8-layer (Llama-2-7B)* & 7.75 & 7.48 & 7.38 & 7.31 & 25.56 \\
    \bottomrule
    \end{tabular}
    \caption{Ablation studies. The default settings are marked with ``*''.}
    \label{tab:ablation}
\end{table*}



\subsection{Ablation Studies}
We perform ablation studies investigate a series of factors which are influential to the performance of Extensible Tokenization, including the down-scaling method, the sampling of scaling factor, and the initialized architecture of the extensible tokenizer. The performances are evaluated with the language modeling task on PG19 and the long-context understanding task on Sing-doc QA (Table \ref{tab:ablation}). 

First of all, we explore the impact of down-scaling with two alternative methods: 1) random down-sampling, which randomly choose $L/k$ embeddings from the extensible tokenizer's output embeddings ($L$ is the chunk size or session length), 2) terminal down-sampling, which select the last $L/k$ results from the extensible tokenizer's output embeddings. Both alternatives are inferior to our default setting, i.e. the strided down-sampling, where the last embedding in every $k$ steps is chosen. In fact, the strided method is not only simple to implement, but also favorable to the representation quality due to the comprehensive coverage of the context window. 

Secondly, we investigate the necessity of dynamically sampling the scaling factor in the training process (denoted as Dynamic Sampling). For comparison, employ a consistent scaling factor throughout the training (denoted as Monotonous). When evaluating the performance of language modeling, both methods utilize the same scaling factor, $k=16$, to extend the context, where the Monotonous setting results in a slightly improved performance. However, when it comes to to single-Doc QA, the extensive tokenizer must work with different scaling factors to accommodate inputs within the context window. In this scenario, dynamic sampling demonstrates a notable advantage over the monotonous method, indicating its versatility to make the extension for diverse context lengths. 

Thirdly, we analyze the impact of the extensible tokenizer's architecture. All alternatives are initialized with LLaMA-2-7B (chat). However, the model sizes differ where the first 4 and 8 transformer layers are taken from the foundation model. It can be observed that the expansion of model size leads to the improved performances on both language modeling and single-Doc QA. This observation is intuitive, as larger models are of higher expressiveness, which is able to make better compression of the context. However, the larger models also lead to extra costs on training and inference. Indeed, it remains to explore the optimal cost-effectiveness of Extensible Tokenization for each specific scenario.

\section{Related Works}

\textbf{Long Context Extension}. Numerous methods have been proposed to extend the context length of LLMs. One important direction involves modifying position encoding, e.g., Position Interpolation \cite{chen2023extending} and NTK-Aware \cite{ntkaware2023}, which allows the LLMs to extend their context lengths during the inference time. These methods can also be applied to the pretrained models and get fine-tuned for better long context generation performance \cite{peng2023yarn}. However, training on extended context data is computationally expensive. Although the training efficiency can be improved by techniques, like LoRA \cite{chen2023longlora,hu2021lora}, sparse attention \cite{chen2023longlora,child2019sparse_transformers}, and FlashAttention \cite{dao2022flashattention}, the cost of training and inference on long context remains substantial. Another line of research focuses on external memory to enhance the LLMs' long context capability. Typically, these methods divide the context by chunks and store them in an additional memory module, which can be retrieved to assist the generation process. For instance, Memorizing Transformers \cite{wu2022memorizing} directly caches key-value pairs of the context, and utilizes the Top-K retrieval to find the most relevant neighbors for the presented query. Similar strategies are adopted by Landmark Attention \cite{mohtashami2023landmark}, which employs landmark tokens to represent the chunks for a better retrieval efficiency, and Focused Transformers \cite{tworkowski2023focused} , which leverages contrastive learning to enhance retrieval accuracy. Finally, the context can also be extended by using sliding windows. For example, StreamingLLM \cite{xiao2023streamingllm} and LM-Infinite \cite{han2023lm_infinite} only maintain the LLM's activations for the latest tokens, which enables the processing of infinite context. Compared with the above methods, Extensible Tokenization has its unique advantages in terms of flexibility, compatibility, and efficiency. Besides, it is able to collaborate with the existing methods for more effective extension of the context. 


\textbf{Context Compression}. There have been continuous effort made for the compression of context. One research direction relies on the explicit compression, where the input text is simplified through summarization or extraction. For instance, LLMLingua \cite{jiang2023llmlingua} introduces a budget controller to maintain semantic integrity under high compression ratios and a token-level iterative compression algorithm to condense the context. RECOMP \cite{xu2023recomp} proposes both extractive and abstractive compressors to compress the context and improve the performance of RAG. Apart from the explicit methods, another research direction focuses on the implicit compression, which are more similar with our work. One early work was made by Funnel-Transformer \cite{dai2020funnel}, which gradually shrinks the sequence length of hidden states in different layers of transformers. However, it calls for the presence of the entire input sequence, which is not appropriate for context extension but mainly for the reduction of computation cost. Besides, several recent works make use of special summarizing tokens to compress the context  \cite{chevalier2023autocompressors,mu2023learning,ge2023context}. In contrast to the previous works, our method is able to bring in superior performances in scaling the LLM's context thanks to its flexible architecture (i.e. the extensive tokenizer with down-scaling) and sample efficient training method (i.e. the two-stream auto-regression). 



\section{Conclusion}
In this paper, we present Extensible Tokenization as a new method to extend the LLM's context. It compresses the raw token embeddings as extensible embeddings, whereby the LLM can perceive more information with the same context window. On top of the auto-regression tasks with the optimized sample efficiency based on two-stream processing, the extensible embeddings can be learned as highly more compact but equally informative representations of the context. Extensible Tokenization is featured by its high flexibility, where the extension for diverse context lengths can be realized by simply making the switch to different scaling factors at the inference time. Besides, the Extensible Tokenization can be introduced as a plug-and-play module, which exhibits a high compatibility with not only the downstream LLM where the extensible tokenizer is trained but also many of its fine-tuned derivatives. Comprehensive experimental studies verify Extensible Tokenization as an effective, efficient, flexible, and compatible method for the extension of LLM's context. 


\clearpage

\bibliographystyle{plain}
\bibliography{ref}

\begin{thebibliography}{10}

\bibitem{ntkaware2023}
Localllama. ntk-aware scaled rope allows llama models to have extended (8k+) con- text size without any fine-tuning and minimal perplexity degradation.
\newblock \url{https://www.reddit.com/r/LocalLLaMA/comments/14lz7j5/ntkaware_scaled_rope_allows_llama_models_to_have/}, 2023.

\bibitem{bai2023longbench}
Yushi Bai, Xin Lv, Jiajie Zhang, Hongchang Lyu, Jiankai Tang, Zhidian Huang, Zhengxiao Du, Xiao Liu, Aohan Zeng, Lei Hou, Yuxiao Dong, Jie Tang, and Juanzi Li.
\newblock Longbench: A bilingual, multitask benchmark for long context understanding.
\newblock {\em arXiv preprint arXiv:2308.14508}, 2023.

\bibitem{beltagy2020longformer}
Iz~Beltagy, Matthew~E. Peters, and Arman Cohan.
\newblock Longformer: The long-document transformer.
\newblock {\em CoRR}, abs/2004.05150, 2020.

\bibitem{aydar2023rmt}
Aydar Bulatov, Yuri Kuratov, and Mikhail~S. Burtsev.
\newblock Scaling transformer to 1m tokens and beyond with {RMT}.
\newblock {\em CoRR}, abs/2304.11062, 2023.

\bibitem{chen2023extending}
Shouyuan Chen, Sherman Wong, Liangjian Chen, and Yuandong Tian.
\newblock Extending context window of large language models via positional interpolation.
\newblock {\em arXiv preprint arXiv:2306.15595}, 2023.

\bibitem{chen2023longlora}
Yukang Chen, Shengju Qian, Haotian Tang, Xin Lai, Zhijian Liu, Song Han, and Jiaya Jia.
\newblock Longlora: Efficient fine-tuning of long-context large language models.
\newblock {\em arXiv preprint arXiv:2309.12307}, 2023.

\bibitem{chevalier2023adapting}
Alexis Chevalier, Alexander Wettig, Anirudh Ajith, and Danqi Chen.
\newblock Adapting language models to compress contexts.
\newblock {\em arXiv preprint 2305.14788}, 2023.

\bibitem{chevalier2023autocompressors}
Alexis Chevalier, Alexander Wettig, Anirudh Ajith, and Danqi Chen.
\newblock Adapting language models to compress contexts.
\newblock In Houda Bouamor, Juan Pino, and Kalika Bali, editors, {\em Proceedings of the 2023 Conference on Empirical Methods in Natural Language Processing, {EMNLP} 2023, Singapore, December 6-10, 2023}, pages 3829--3846. Association for Computational Linguistics, 2023.

\bibitem{child2019sparse_transformers}
Rewon Child, Scott Gray, Alec Radford, and Ilya Sutskever.
\newblock Generating long sequences with sparse transformers.
\newblock {\em arXiv preprint arXiv:1904.10509}, 2019.

\bibitem{together2023redpajama}
Together Computer.
\newblock Redpajama: an open dataset for training large language models, 2023.

\bibitem{longchat2023}
Li~Dacheng, Shao Rulin, Xie Anze, Sheng Ying, Zheng Lianmin, E.~Gonzalez Joseph, Stoica Ion, Ma~Xuezhe, and Zhang Hao.
\newblock How long can open-source llms truly promise on context length?, June 2023.

\bibitem{dai2020funnel}
Zihang Dai, Guokun Lai, Yiming Yang, and Quoc Le.
\newblock Funnel-transformer: Filtering out sequential redundancy for efficient language processing.
\newblock In Hugo Larochelle, Marc'Aurelio Ranzato, Raia Hadsell, Maria{-}Florina Balcan, and Hsuan{-}Tien Lin, editors, {\em Advances in Neural Information Processing Systems 33: Annual Conference on Neural Information Processing Systems 2020, NeurIPS 2020, December 6-12, 2020, virtual}, 2020.

\bibitem{dao2023flash}
Tri Dao.
\newblock Flashattention-2: Faster attention with better parallelism and work partitioning.
\newblock {\em CoRR}, abs/2307.08691, 2023.

\bibitem{dao2022flashattention}
Tri Dao, Dan Fu, Stefano Ermon, Atri Rudra, and Christopher R{\'e}.
\newblock Flashattention: Fast and memory-efficient exact attention with io-awareness.
\newblock {\em Advances in Neural Information Processing Systems}, 35:16344--16359, 2022.

\bibitem{ding2023longnet}
Jiayu Ding, Shuming Ma, Li~Dong, Xingxing Zhang, Shaohan Huang, Wenhui Wang, Nanning Zheng, and Furu Wei.
\newblock Longnet: Scaling transformers to 1,000,000,000 tokens.
\newblock {\em arXiv preprint arXiv:2307.02486}, 2023.

\bibitem{pile}
Leo Gao, Stella Biderman, Sid Black, Laurence Golding, Travis Hoppe, Charles Foster, Jason Phang, Horace He, Anish Thite, Noa Nabeshima, Shawn Presser, and Connor Leahy.
\newblock The {P}ile: An 800gb dataset of diverse text for language modeling.
\newblock {\em arXiv preprint arXiv:2101.00027}, 2020.

\bibitem{ge2023context}
Tao Ge, Jing Hu, Xun Wang, Si-Qing Chen, and Furu Wei.
\newblock In-context autoencoder for context compression in a large language model.
\newblock {\em arXiv preprint arXiv:2307.06945}, 2023.

\bibitem{han2023lm_infinite}
Chi Han, Qifan Wang, Wenhan Xiong, Yu~Chen, Heng Ji, and Sinong Wang.
\newblock Lm-infinite: Simple on-the-fly length generalization for large language models.
\newblock {\em CoRR}, abs/2308.16137, 2023.

\bibitem{hu2021lora}
Edward~J Hu, Yelong Shen, Phillip Wallis, Zeyuan Allen-Zhu, Yuanzhi Li, Shean Wang, Lu~Wang, and Weizhu Chen.
\newblock Lora: Low-rank adaptation of large language models.
\newblock {\em arXiv preprint arXiv:2106.09685}, 2021.

\bibitem{huang2023selective_cache}
Xinting Huang and Nora Hollenstein.
\newblock Long-range language modeling with selective cache.
\newblock In Houda Bouamor, Juan Pino, and Kalika Bali, editors, {\em Findings of the Association for Computational Linguistics: {EMNLP} 2023, Singapore, December 6-10, 2023}, pages 4838--4858. Association for Computational Linguistics, 2023.

\bibitem{jiang2023llmlingua}
Huiqiang Jiang, Qianhui Wu, Chin-Yew Lin, Yuqing Yang, and Lili Qiu.
\newblock Llmlingua: Compressing prompts for accelerated inference of large language models.
\newblock {\em arXiv preprint arXiv:2310.05736}, 2023.

\bibitem{mohtashami2023landmark}
Amirkeivan Mohtashami and Martin Jaggi.
\newblock Landmark attention: Random-access infinite context length for transformers.
\newblock {\em arXiv preprint arXiv:2305.16300}, 2023.

\bibitem{mu2023learning}
Jesse Mu, Xiang~Lisa Li, and Noah Goodman.
\newblock Learning to compress prompts with gist tokens.
\newblock {\em arXiv preprint arXiv:2304.08467}, 2023.

\bibitem{wu2023gist}
Jesse Mu, Xiang~Lisa Li, and Noah~D. Goodman.
\newblock Learning to compress prompts with gist tokens.
\newblock {\em CoRR}, abs/2304.08467, 2023.

\bibitem{peng2023yarn}
Bowen Peng, Jeffrey Quesnelle, Honglu Fan, and Enrico Shippole.
\newblock Yarn: Efficient context window extension of large language models.
\newblock {\em arXiv preprint arXiv:2309.00071}, 2023.

\bibitem{raecompressive2019}
Jack~W Rae, Anna Potapenko, Siddhant~M Jayakumar, Chloe Hillier, and Timothy~P Lillicrap.
\newblock Compressive transformers for long-range sequence modelling.
\newblock {\em arXiv preprint}, 2019.

\bibitem{rae2020compressive}
Jack~W. Rae, Anna Potapenko, Siddhant~M. Jayakumar, Chloe Hillier, and Timothy~P. Lillicrap.
\newblock Compressive transformers for long-range sequence modelling.
\newblock In {\em 8th International Conference on Learning Representations, {ICLR} 2020, Addis Ababa, Ethiopia, April 26-30, 2020}. OpenReview.net, 2020.

\bibitem{roziere2023code}
Baptiste Roziere, Jonas Gehring, Fabian Gloeckle, Sten Sootla, Itai Gat, Xiaoqing~Ellen Tan, Yossi Adi, Jingyu Liu, Tal Remez, J{\'e}r{\'e}my Rapin, et~al.
\newblock Code llama: Open foundation models for code.
\newblock {\em arXiv preprint arXiv:2308.12950}, 2023.

\bibitem{touvron2023llama}
Hugo Touvron, Thibaut Lavril, Gautier Izacard, Xavier Martinet, Marie-Anne Lachaux, Timoth{\'e}e Lacroix, Baptiste Rozi{\`e}re, Naman Goyal, Eric Hambro, Faisal Azhar, et~al.
\newblock Llama: Open and efficient foundation language models.
\newblock {\em arXiv preprint arXiv:2302.13971}, 2023.

\bibitem{touvron2023llama-b}
Hugo Touvron, Louis Martin, Kevin Stone, Peter Albert, Amjad Almahairi, Yasmine Babaei, Nikolay Bashlykov, Soumya Batra, Prajjwal Bhargava, Shruti Bhosale, et~al.
\newblock Llama 2: Open foundation and fine-tuned chat models.
\newblock {\em arXiv preprint arXiv:2307.09288}, 2023.

\bibitem{tworkowski2023focused}
Szymon Tworkowski, Konrad Staniszewski, Miko{\l}aj Pacek, Yuhuai Wu, Henryk Michalewski, and Piotr Mi{\l}o{\'s}.
\newblock Focused transformer: Contrastive training for context scaling.
\newblock {\em arXiv preprint arXiv:2307.03170}, 2023.

\bibitem{wu2022memorizing}
Yuhuai Wu, Markus~Norman Rabe, DeLesley Hutchins, and Christian Szegedy.
\newblock Memorizing transformers.
\newblock In {\em The Tenth International Conference on Learning Representations, {ICLR} 2022, Virtual Event, April 25-29, 2022}. OpenReview.net, 2022.

\bibitem{xiao2023streamingllm}
Guangxuan Xiao, Yuandong Tian, Beidi Chen, Song Han, and Mike Lewis.
\newblock Efficient streaming language models with attention sinks.
\newblock {\em arXiv preprint arXiv:2309.17453}, 2023.

\bibitem{xiao2023efficient}
Guangxuan Xiao, Yuandong Tian, Beidi Chen, Song Han, and Mike Lewis.
\newblock Efficient streaming language models with attention sinks.
\newblock {\em arXiv preprint arXiv:2309.17453}, 2023.

\bibitem{xu2023recomp}
Fangyuan Xu, Weijia Shi, and Eunsol Choi.
\newblock Recomp: Improving retrieval-augmented lms with compression and selective augmentation.
\newblock {\em arXiv preprint arXiv:2310.04408}, 2023.

\bibitem{zaheer2020bigbird}
Manzil Zaheer, Guru Guruganesh, Kumar~Avinava Dubey, Joshua Ainslie, Chris Alberti, Santiago Ontanon, Philip Pham, Anirudh Ravula, Qifan Wang, Li~Yang, et~al.
\newblock Big bird: Transformers for longer sequences.
\newblock {\em Advances in neural information processing systems}, 33:17283--17297, 2020.

\end{thebibliography}

\end{document}